\documentclass[letterpaper, 10 pt, conference]{ieeeconf}  % Comment this line out if you need a4paper

\IEEEoverridecommandlockouts                              % This command is only needed if 
\usepackage{graphics} % for pdf, bitmapped graphics files
\usepackage{epsfig} % for postscript graphics files
\usepackage{booktabs} % For professional-looking tables (toprule, midrule, bottomrule)
\usepackage{siunitx}  % For scientific notation and number formatting
\usepackage{multirow} % For multi-row cells if needed
\usepackage{array}    % For better column definitions
\usepackage{makecell}
\usepackage[style=numeric, sorting=none]{biblatex}
\usepackage{hyperref} 
\usepackage{amsfonts}
\usepackage{listings}
\usepackage{xcolor}
\usepackage{amsmath}  % For advanced math formatting
\usepackage{amssymb}  % For symbols like \mathbb{1}
\usepackage{bbm} 
\usepackage{newtxtext, newtxmath} % Alternative for \mathbb{1} (indicator function)
\usepackage{cleveref}
\usepackage{balance}   % in preamble

\usepackage{subfig}

\definecolor{codegreen}{rgb}{0,0.6,0}
\definecolor{codegray}{rgb}{0.5,0.5,0.5}
\definecolor{codepurple}{rgb}{0.58,0,0.82}
\definecolor{backcolour}{rgb}{0.95,0.95,0.92}

\lstdefinestyle{mystyle}{
    backgroundcolor=\color{backcolour},   
    commentstyle=\color{codegreen},
    keywordstyle=\color{magenta},
    numberstyle=\tiny\color{codegray},
    stringstyle=\color{codepurple},
    basicstyle=\ttfamily\scriptsize,
    breakatwhitespace=false,         
    breaklines=true,                 
    captionpos=b,                    
    keepspaces=true,                 
    numbers=left,                    
    numbersep=5pt,                  
    showspaces=false,                
    showstringspaces=false,
    showtabs=false,                  
    tabsize=2
}

\title{\LARGE \bf
Scaling Multi-Agent Reinforcement Learning for Underwater Acoustic Tracking via Autonomous Vehicles
}

\author{Matteo Gallici, Ivan Masmitja, and Mario Martín% <-this % stops a space
\thanks{M. G. is with the KEMLG Research Group, Universitat Politècnica de Catalunya Barcelona, Spain. \texttt{gallici@cs.upc.edu}}%
\thanks{I. M. is with the Instituto de Ciencias del Mar, Consejo Superior de Investigaciones Científicas, Barcelona, Spain. \texttt{masmitja@icm.csic.es}}%
\thanks{M. M. is with the KEMLG Research Group, Universitat Politècnica de Catalunya (UPC), and with the HPAI group at Barcelona Supercomputing Center (BSC),  Barcelona, Spain. \texttt{mmartin@cs.upc.edu}}%
}

\begin{document}

\maketitle
%\thispagestyle{firstpage}
%\pagestyle{empty}

%%%%%%%%%%%%%%%%%%%%%%%%%%%%%%%%%%%%%%%%%%%%%%%%%%%%%%%%%%%%%%%%%%%%%%%%%%%%%%%%
\begin{abstract}

Autonomous vehicles (AVs) offer a cost-effective solution for scientific missions such as underwater tracking. Reinforcement learning (RL) has emerged as a powerful method for controlling AVs, but scaling to fleets (essential for multi-target tracking or rapidly moving targets) is challenging. Multi-Agent RL (MARL) is notoriously sample-inefficient, and while high-fidelity simulators like Gazebo’s LRAUV provide up to 100× faster-than-real-time single-robot simulations, they offer little speedup in multi-vehicle scenarios, making MARL training impractical. Yet, high-fidelity simulation is crucial to test complex policies and close the sim-to-real gap. To address these limitations, we develop a GPU-accelerated environment that achieves up to 30,000× speedup over Gazebo while preserving its dynamics. This enables fast, end-to-end GPU training and seamless transfer to Gazebo for evaluation. We also introduce a Transformer-based architecture (\textit{TransfMAPPO}) that learns policies invariant to fleet size and number of targets, enabling curriculum learning to train larger fleets on increasingly complex scenarios. After large-scale GPU training, we perform extensive evaluations in Gazebo, showing our method maintains tracking errors below 5m even with multiple fast-moving targets.
\end{abstract}

%%%%%%%%%%%%%%%%%%%%%%%%%%%%%%%%%%%%%%%%%%%%%%%%%%%%%%%%%%%%%%%%%%%%%%%%%%%%%%%%
\section{INTRODUCTION}

Underwater tracking (UT) is essential to advance marine research missions, such as tracking marine species and oceanographic phenomena \cite{zarokanellos2022frontal, masmitja2020mobile}, monitoring large-scale data collection using autonomous underwater vehicles (AUV) \cite{Yanwu2021}, and managing marine protected areas \cite{vigo2021spatial}. However, underwater communication systems face challenges such as low reliability and bandwidth, as GPS is ineffective \cite{heidemann2012underwater}. Acoustic tracking using AUVs or autonomous surface vehicles (ASVs) offers a promising solution, improving efficiency and reducing costs compared to traditional fixed equipment \cite{masmitja2020mobile, Yanwu2021}. Yet, challenges like unreliable communication, complex environment dynamics, and energy constraints necessitate advanced motion planning to enhance AVs' tracking capabilities.

   \begin{figure}[thpb]
      \centering
      \includegraphics[scale=0.65]{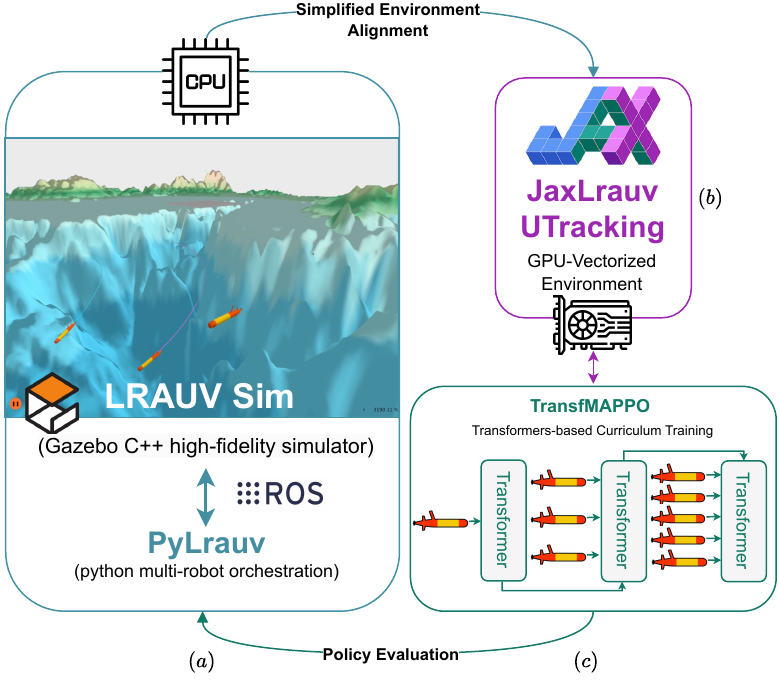}
      \caption{\small Overview of our training and evaluation pipeline. \textbf{PyLrauv} $(a)$ is a new Python package to control multiple robots in the C++ high-fidelity LRAUV simulator. \textbf{JaxLrauv} $(b)$ is a GPU-accelerated, simplified environment that supports massive parallelization while preserving the dynamics of the LRAUV simulator. \textbf{TransfMAPPO} $(c)$ employs transformers to train progressively larger fleets of vehicles to coordinate via curriculum learning. The final policies trained in JaxLrauv with TransfMAPPO are then evaluated in the realistic LRAUV simulator prior to real-world deployment.}
      \label{fig:env-distillation}
   \end{figure}

\begin{figure}[thpb]
    \includegraphics[scale=0.48]{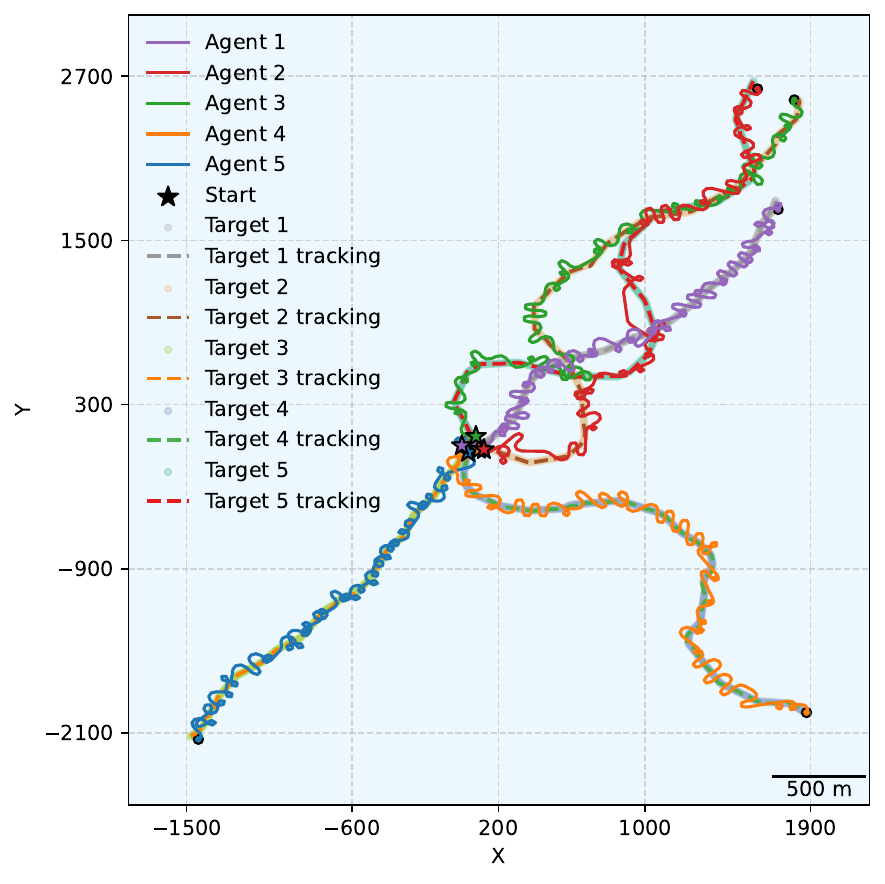}
    \caption{\small Five agents trained in the GPU simplified environment follow in Gazebo simulator five fast targets over several kms. \href{https://mttga.github.io/posts/pylrauv/images/5v5.gif}{Video: https://mttga.github.io/posts/pylrauv/images/5v5.gif}}
    \label{fig:traj_5v5}
\end{figure}

Reinforcement learning (RL) enables AVs to learn optimal navigation strategies through trial and error, providing dynamic responses to environmental conditions compared to pre-programmed approaches \cite{cote2019characterizing}. Preliminary work by \cite{ivanscience} demonstrated the potential of using RL to train an ASV policy for tracking an underwater target using range-only acoustic data. More recent studies \cite{multiuav-pro, multi-uav-simple} have shown that Multi-Agent Reinforcement Learning (MARL) can enable cooperative tracking by sharing data collected simultaneously from different locations. However, these MARL studies were conducted in very simple, abstract environments. Deploying such a complex system in the sea requires extensive preliminary testing. Simulators such as Gazebo's Long-Range AUV simulator (LRAUV Sim)~\cite{lrauv_sim_icra}, which models hydrodynamics, acoustic communication, and marine sensors of real vehicles, are crucial tools for this phase. Unfortunately, LRAUV Sim provides only about a $10\times$ faster-than-real-time speedup in multi-robot simulations, and its C++-based control complicates integration with RL methods, making direct application of MARL impractical in this simulator.

To advance multi-agent systems for underwater tracking, we therefore make the following contributions:

\begin{itemize}
\item \textbf{Fast GPU Training and High-Fidelity Testing Pipeline}: We propose a new pipeline for scaling MARL in underwater tracking scenarios, composed of two interconnected frameworks (see \autoref{fig:env-distillation}). The first, \textit{PyLrauv}, is an open-source Python package built on ROS2 \cite{ros2} that provides a Gym-like interface for controlling LRAUV Sim. PyLrauv enables seamless control and observation of a variable number of agents and targets in the Gazebo high-fidelity backend, and also supports direct deployment on real vehicles. The second, \textit{JaxLrauv}, is a simplified environment implemented on top of JaxMARL \cite{flair2023jaxmarl}, a popular GPU-accelerated MARL framework. JaxLrauv achieves up to a 30,000$\times$ speedup over Gazebo, enabling training of effective policies in minutes, while preserving full compatibility and transferability with Gazebo and real vehicles through direct equivalence with the PyLrauv controller.

\item \textbf{Agent-Target Invariant Policies with Transformers and Curriculum Learning}: By integrating Transformers into a novel variant of MAPPO~\cite{mappo} (\textit{TransfMAPPO}), we train policies that are invariant to the number of agents and targets. This enables a curriculum learning (CL) pipeline that progressively trains larger fleets in increasingly complex multi-target scenarios. TransfMAPPO learns robust cooperative policies where traditional from-scratch approaches such as MAPPO fail. Thanks to the combination of fast training and curriculum learning, we obtain a general policy capable of tracking up to 5 simultaneous targets with only 5 vehicles (f.i. \autoref{fig:traj_5v5}), whereas state-of-the-art require up to 12 vehicles to track 4 targets.
\end{itemize}

\section{RELATED WORK}

\textcolor{black}{Early research on underwater tracking (UT) with autonomous vehicles has primarily relied on traditional control strategies for AUVs~\cite{heidemann2012underwater, almeida2017}. A promising new direction was introduced by~\cite{ivanscience}, who incorporated RL to improve navigation and tracking robustness. However, their approach was limited to a single vehicle tracking fixed or slowly moving targets. More recent studies~\cite{multi-uav-simple, multiuav-pro} have explored the use of MARL to train cooperative policies for tracking moving targets. These methods, however, were evaluated in highly abstract and simplified environments. Our proposed pipeline, by contrast, is designed with implementation considerations in mind and aims to ensure that learned policies remain robust in more realistic scenarios. In particular,\cite{multi-uav-simple} demonstrated cooperative behaviour for a fixed configuration of two vehicles tracking a single target. Similarly to our approach,\cite{multiuav-pro} employed an attention mechanism. However, their policies were trained separately for fixed team configurations (e.g., 4 agents tracking 2 targets, 6 agents tracking 3 targets, or 12 agents tracking 4 targets), and did not aim to achieve invariance with respect to team size. In contrast, our method explicitly targets generalisation across varying numbers of agents and targets, and we demonstrate robust tracking of 5 targets using 5 vehicles within a single scalable framework. Neither~\cite{multi-uav-simple} nor~\cite{multiuav-pro} report wall-clock training times or provide official codebases, which makes a direct comparison of computational efficiency infeasible. More importantly, their primary objective differs from ours: they focus on learning coordination policies for fixed team sizes, whereas our work emphasises scalability and curriculum-based training across increasing team sizes.}

Our TransfMAPPO architecture is inspired by TransfQMix~\cite{transfqmix}, but adopts PPO instead of Q-Learning as the RL backbone due to its natural compatibility with parallelized environments. Moreover, TransfMAPPO replaces the centralized hypernetwork with a centralized critic, simplifying the model structure. %Unlike TransfQMix, TransfMAPPO also supports continuous action spaces, rather than being restricted to discrete ones. 
Several other methods have explored scaling coordination in MARL using attention mechanisms, Graph Neural Networks (GNNs), or Transformers~\cite{nayak2023scalable, goeckner2024graph, cai2024transformer, updet}, but a detailed comparison with these approaches is beyond the scope of this work. The most similar work to ours that combines transformers with MAPPO is Multi-Agent Transformer~\cite{wen2022multi}. However, in that work transformers are used to model the multi-agent problem as a sequential decision process, whereas in our case we employ them to improve simultaneous coordination. Related to our approach,~\cite{zhang2022automatic, wang2020few} employ curriculum learning to progressively train policies that adapt to larger team sizes, but these methods have not been evaluated in complex robotic scenarios.

% Frameworks like JaxMARL \cite{flair2023jaxmarl} have demonstrated the benefits of GPU-accelerated MARL, albeit in relatively simple environments, while packages such as IsaacGym \cite{isaacgym} provide more advanced GPU-based simulation of real robots but do not offer the hydrodynamic models and communication protocols necessary to bridge the sim-to-real gap.

% Our pipeline can be associated with model-based RL \cite{model_based_1, model_based_2, model_based_3}, which aim to learn a world model to improve and speed up RL training. Our method creates a simplified, GPU-accelerated physical model from a high-fidelity simulator, but instead of focusing on step-wise alignment between the world model and the real world, it focuses on preserving high-level dynamics at a different time scale with respect to a more complex simulator. Moreover, our distilled environment is not accessible by our learning algorithm, as is usually the case in model-based RL; instead, we simply use \textit{model-free} RL on top of it.

\section{BACKGROUND}

\subsection{Underwater Localization with Autonomous Vehicles}

Underwater acoustic tracking with autonomous vehicles is a member of the family of Range Only Single Beacon (ROSB) localization techniques \cite{rosb, av_range_only}, which use range-only acoustic observations gathered over time to locate a target. For static targets, the trilateration problem can be linearised by least squares (LS) algorithms, which assume static conditions and incur errors when targets move \cite{ThesisLisboa}. For dynamic scenarios, particle filters (PF) use Bayesian estimation, representing potential target states with weighted particles updated via motion models and resampled based on measurement likelihoods \cite{ROSB1, ROSB2}.

\subsection{Multi-Agent Reinforcement Learning}
Cooperative multi-agent tasks are formalized as decentralized partially observable Markov decision processes (Dec-POMDP), %\cite{dec-Pompd}
defined by the tuple \( G = \langle S, \mathbf{U}, P, r, Z, O, n, \gamma \rangle \), where \( S \) is the global state space, \( \mathbf{U} = U^n \) is the joint action space for \( n \) agents (each selecting actions \( u^a \in U \)), \( P(s' \mid s, \mathbf{u}) \) is the state transition function, \( r(s, \mathbf{u}) \) is the shared reward function, \( Z \) is the observation space for each agent, \( O(s, a) \) defines the observation \( z^a \in Z \) for agent \( a \), and \( \gamma \in [0,1) \) is the discount factor. At each timestep, agent $a$ receives a partial observation $z^a = O(s, a)$ and maintains an action-observation history $\tau^a = (z_0^a, u_0^a, \dots, z_t^a) \in \mathcal{T}$. The agent’s policy $\pi^a(u^a \mid \tau^a): \mathcal{T} \times U \rightarrow [0,1]$ maps this history to a distribution over actions. The agents collectively aim to maximize the expected discounted return $\mathbb{E}[\sum_{t=0}^\infty \gamma^t r_t]$. A key paradigm in MARL is centralized training with decentralized execution (CTDE) \cite{rashid2018qmixmonotonicvaluefunction}, where agents leverage centralized information during training but act independently during execution. This allows the use of a centralized critic that estimates the value function $V(S)$ using the global state $S$, while policies $\pi^a$ remain decentralized, depending only on local histories $\tau^a$.

MAPPO \cite{mappo} extends Proximal Policy Optimization (PPO) \cite{ppo} to multi-agent settings. Each agent’s policy (actor) $\pi_\theta^a$ and centralized value function (critic) $V_\phi$ are optimized jointly. The actors are updated using the standard PPO clipped surrogate objective, while a single critic estimates the value function of the global environment's state. During execution, only the actors are used, enabling decentralized control.

\section{METHOD}
\label{sec:method}

\subsection{Fast GPU training to High-fidelity testing Pipeline}
\subsubsection{PyLrauv}
\label{sec:environment}

PyLrauv provides a Python interface to LRAUV Sim by leveraging ROS 2 \cite{ros2}. This is achieved by integrating a set of additional communication handlers into the official Gazebo C++ library, \texttt{ros\_gz}\footnote{\href{https://github.com/gazebosim/ros_gz}{https://github.com/gazebosim/ros\_gz}}. The integration allows control of the LRAUV robots with Python using \texttt{rclpy} \footnote{\href{https://github.com/ros2/rclpy}{https://github.com/ros2/rclpy}}. We have developed a set of Python controllers to send commands to the LRAUV vehicles in Gazebo and observe their published states. Since the actual LRAUV vehicles use the same technology, the controllers provided within PyLrauv can also be used to control real LRAUV robots. 

%Controlling multiple vehicles simulated in Gazebo in a Gym-like fashion is straightforward, as demonstrated by the following code:

% \begin{lstlisting}[language=Python]
% from pylrauv import UTrackingEnv

% env = UTrackingEnv(num_agents=2, num_targets=2)
% obs, state = env.reset()
% actions = {'agent_1': 0, 'agent_2': 0}
% obs, state, reward, done = env.step(actions, step_time=30)
% \end{lstlisting}

\subsubsection{PyLrauv UTracking Environment}
\texttt{UTrackingEnv} is a Gym-like environment managing agent and target commands, communication, tracking, and RL tasks like observation building and reward calculations. Agents communicate via the Gazebo \texttt{LRAUV Acoustic Comms Plugin} and collect target ranges using the \texttt{LRAUV Range Bearing Plugin}. At the start of an episode, agents are spawned on the sea surface and targets at random depths, ensuring a minimum distance between each entity of 50 m and maximum of 200 m. Both agents and targets are LRAUV vehicles. Each step lasts 30 seconds in our experiments, at the beginning of which agents listen for range signals from targets, and at the end broadcast their locations and observations to other agents (communication phase). Each agent maintains its own tracking model (LS or PF) for each target, updated with all the information received. This ensures a genuine decentralization of the system. Targets are tracked in 2D space, as depth is typically known \cite{ivanscience}.

\paragraph{Observation Space}
\label{par:observation}
The agents receive their partial observation of the environment at the end of the step phase, which includes the relative distances to all other entities in the 3D space. For other agents, these distances are calculated using the information they have received from them (if available), and for targets, the distances are calculated from the tracking predictions generated by their own tracking modules. \cite{ivanscience} incorporates the range observations into observation vectors, which may be helpful, but may also cause agents to overfit to training dynamics and overlook tracking information. To prevent this, we limit observations to tracking-derived values, ensuring agents rely on tracking data and its inherent errors. The environment also returns the global state of the environment, which includes the true 3D position of each entity, its velocity, and its direction.

\paragraph{Reward Function}
We define two possible reward functions. The \textbf{tracking} reward is designed to reduce the global tracking error. Let $e_i$ denote the tracking error for target $i$, $\epsilon_{\text{min}}$ the ideal error of the system, n$\epsilon_{\text{max}}$ the maximum rewarding error, and define the reward for that target as an exponential decay function:

\begin{equation}
\small
r_i^{tracking} = \begin{cases}
1, & e_i < \epsilon_{\text{min}}, \\
\exp\left(-\dfrac{2t}{1-t}\right), & \epsilon_{\text{min}} \le e_i \le \epsilon_{\text{max}}, \\
0, & e_i > \epsilon_{\text{max}},
\end{cases}
\end{equation}
with $t = (e_i - \epsilon_{\text{min}})/({\epsilon_{\text{max}} - \epsilon_{\text{min}}})$. We set $\epsilon_{\text{min}} = 10\,\text{m}$ and $\epsilon_{\text{max}} = 50\,\text{m}$. The exponential decay encourages precise tracking, distinguishing more clearly states where the tracking was closer to the ideal tracking error. 
The \textbf{follow} reward is based on the proper distribution of agents with respect to the targets. This reward encourages a perfect distribution across targets for the agents, which is known to be the best policy when $N$ agents need to follow (rapid enough) $N$ targets. For target $i$, if the minimum distance between it and any agent, $d_i$, is less than specified threshold, $d_{\min}$, then the target is considered successfully followed. The reward is computed as
\begin{equation}
r_i^{follow} = \mathbb{I}\left\{ d_i \leq \,d_{\min} \right\},
\end{equation}
with the threshold set in our experiments to $50\,\text{m}$ (and $100\,\text{m}$ for fast targets). In both cases, the global reward to the system is given by
$r = \frac{1}{N}\sum_{i=1}^{N} r_i,$
where $N$ is the number of targets to track, i.e. the reward is normalized by the number of targets to be always in the interval $[0,1]$, so that learned value functions can be transferred across scenarios with different numbers of targets. In addition, a crash penalty is always applied if the distance between any two agents falls below a minimum valid distance, $d_{safe}$, i.e., $
r = -1 \quad \text{if} \quad \min_{i \neq j}\{d_{ij}\} < d_{safe}.$

\paragraph{Actions Space} \textcolor{black}{We keep the agents’ velocity constant and allow them to control only the rudder. This simplifies the action space of the multi-agent system and is equivalent to controlling the vehicle’s heading. This approach is robust and can be transferred to real robots by using low-level velocity and rudder controllers to ensure that the vehicle reaches a specified heading. The rudder is discretised into five values corresponding to equidistant angles ranging from -0.24 to 0.24 radians. This is equivalent to a heading change between -1.45 and 1.45 radians when the velocity is 1 m/s and the time step is 30 seconds, as in our experiments. Agents can adjust the rudder only to one of the two adjacent angles, thereby preventing abrupt changes in direction.}

\paragraph{Termination} The episode ends when the maximum number of steps is reached. % The environment is not truncated when crashes occur or targets are lost, as we experimentally observed that truncating episodes would create training instabilities. 

\textbf{Why is UTracking a difficult environment?}  
UTracking poses a challenging benchmark due to partial observability, where agents can only detect targets within 450\,m using noisy acoustic signals and share data through intermittent communication limited to 1500\,m. The environment is highly stochastic, with sensor noise, communication dropouts, environmental perturbations, and unpredictable target motions. Effective tracking requires near-perfect coordination among agents, especially when following multiple fast-moving targets, as poor coordination can quickly lead to target loss. Finally, the long horizons of hour-scale missions (thousands of steps) demand learning policies that handle delayed consequences and remain robust to error accumulation in non-stationary multi-agent settings.

\subsubsection{JaxLrauv UTracking}

LRAUV Sim simulates the dynamics of an LRAUV robot at the millisecond level. However, in practice, LRAUV vehicles take actions only every few seconds to minutes, while the tracking missions of interest span several hours. For this reason, it is beneficial to simulate the environment state at a much larger time scale. Formally, we are interested in modeling the new position in the 2D plane $\mathbf{p}_{t+1} \in \mathbb{R}^{2}$ after a time step $\delta_t$, given the current robot position $\mathbf{p}_t$, absolute speed $v$, and rudder angle $\gamma$. Instead of re-building an accurate physical model (already provided by Gazebo), we approximate the change in the robot’s heading, $\delta_\psi$, in order to use a simplified trajectory model:  $\mathbf{p}_{t+1} = \mathbf{p}_t + v\delta_t \begin{bmatrix} \cos(\psi_t + \delta_\psi), \sin(\psi_t + \delta_\psi) \end{bmatrix}$. Fixing $v$ and $\delta_t$, we parametrize $\delta_\psi$ with respect to the rudder angle through a function $\theta(\gamma)$. To approximate $\theta(\gamma)$, we collect trajectories in PyLrauv and train a supervised model that predicts $\delta_\psi$ from $\theta(\gamma)$. We found that using a linear model for each fixed combination of $v$ and $\delta_t$, $\theta(\gamma)$ can be approximated with a mean absolute error below 0.015 radians and a global $R^2$ score of 0.99. Based on this, we build an ensemble of linear models for different combinations of $v$ and $\delta_t$. While such a simple model cannot capture long-term dependencies and diverges when used autoregressively for long trajectories, it performs well for shorter trajectories and is much faster than, for example, an LSTM or a handwritten physical system. To improve robustness, we introduce Gaussian noise with a standard deviation of 0.02 radians, forcing the trained agent to handle trajectory uncertainty. In addition, we equip the agents with recurrent mechanisms. We found that training in this simplified environment leads to policies that transfer directly to Gazebo.

The rest of the environment, including measurement noise, communication loss, partial observability, and collision handling, is implemented manually based on data collected in Gazebo. The final simplified environment is implemented in JAX, enabling an end-to-end GPU-based RL pipeline. To further improve training efficiency, we implement a Particle Filter in JAX, allowing the tracking phase to also benefit from GPU acceleration. By nesting vectorized functions, we update all particles across targets, agents, and parallel environments in a single pass, achieving both high speed and precise tracking during training. The same vectorized PF algorithm is also integrated into PyLrauv for consistency. The speedups achieved across different agent–target configurations are reported in \autoref{tab:speedup}. These gains make it possible to train a single-target tracking model that can be directly deployed in LRAUV Sim in as little as 10 minutes.

\begin{table}[h]
\caption{ \small Speedups Obtained with the Accelerated Environment. Steps Per Second (SPS) and relative speedup.}
\centering
\resizebox{\columnwidth}{!}{%
\begin{tabular}{l c c c c c c c}
\toprule
Config & \multicolumn{1}{c}{Pylrauv} & \multicolumn{2}{c}{JaxLrauv 1 Envs} & \multicolumn{2}{c}{JaxLrauv 128 Envs} & \multicolumn{2}{c}{JaxLrauv 1024 Envs} \\
\cmidrule(lr){2-2} \cmidrule(lr){3-4} \cmidrule(lr){5-6} \cmidrule(lr){7-8}
       & SPS   & SPS   & Speedup  & SPS    & Speedup  & SPS    & Speedup  \\
\midrule
1A,1T  & 2.7   & 1289  & 477      & 68508  & 25352   & 81686  & 30229    \\
2A,2T  & 1.0   & 1068  & 1084     & 19712  & 20017   & 21534  & 21867    \\
3A,3T  & 0.4   & 1053  & 2439     & 9128   & 21139   & 9786   & 22663    \\
4A,4T  & 0.3   & 1020  & 3571     & 5180   & 18125   & 5450   & 19070    \\
5A,5T  & 0.2   & 936   & 4751     & 3469   & 17607   & 3574   & 18142    \\
\bottomrule
\label{tab:speedup}
\end{tabular}%
}
\end{table}
\vspace{-2em}

% \subsection{Vectorized Particle Filter}

% While PF represent a state-of-the-art method for underwater tracking, their computational expense renders them inefficient for training, particularly in multi-agent scenarios. To mitigate this, \cite{ivanscience} suggests employing Least Squares (LS) during training and reserving PF for testing. However, LS is only stable for static or slow-moving targets, limiting its scalability for faster targets and potentially intensifying the inherent non-stationarity of MARL. To facilitate efficient training, we implement a PF in JAX, utilizing \texttt{vmap} for GPU parallelization. By nesting vmapped functions, during training we update all particles across targets, agents, and parallel environments in a single pass, reducing the speed loss to only 30\% compared to LS. As demonstrated in \autoref{fig:pf_vs_ls}, training with LS fails for targets moving faster than one-third of the agent's velocity, whereas PF maintains stability.

% \begin{figure}[thpb]
%       \centering
%       \includegraphics[scale=0.4]{images/training/tracking_type_training.pdf}
%       \caption{Using Particle Filter or Least Squares to train an agent to follow a target at different velocities.}
%       \label{fig:pf_vs_ls}
%       \vspace{-1em}
% \end{figure}

\subsection{Curriculum Learning Via Transformers}
\subsubsection{TransfMAPPO}

\begin{figure}[thpb]
      \centering
      \includegraphics[scale=0.35]{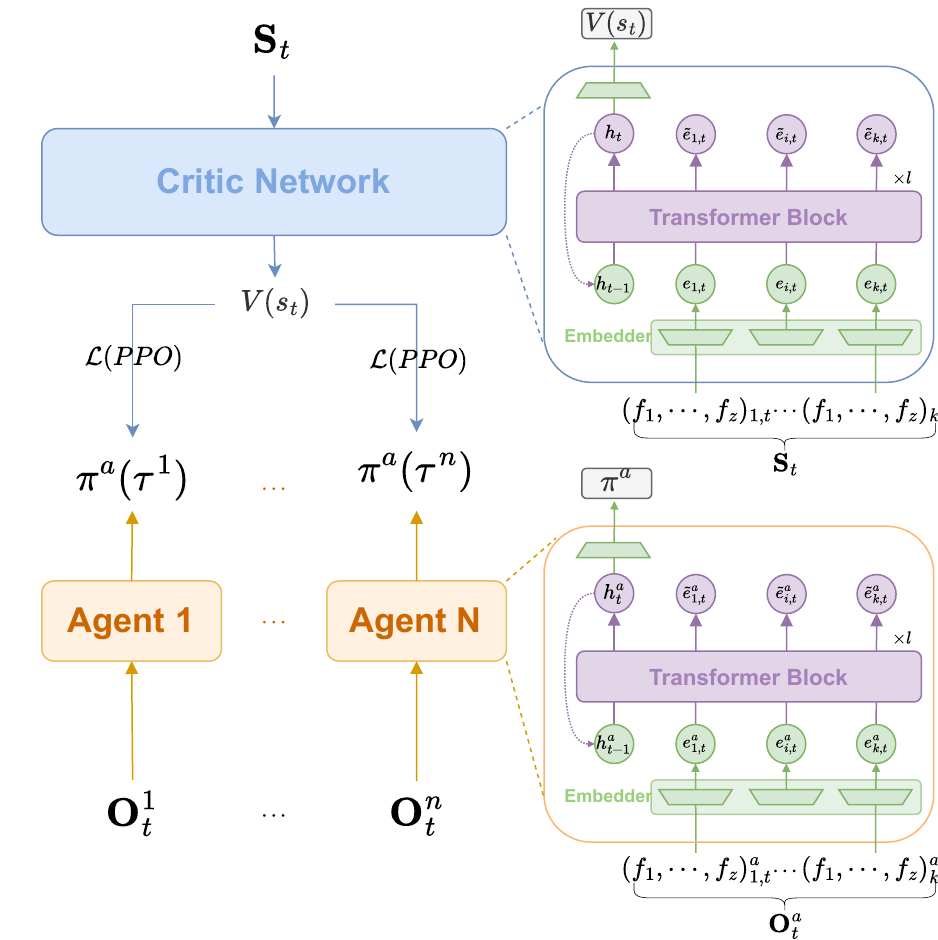}
      \caption{\small \textcolor{black}{In TransfMAPPO, both each agent (actor) and the centralized critic are implemented as Transformer networks. Each TransformerAgent observes the environment as a set of entity vectors describing itself, its teammates, and the targets. Each vector encodes the (possibly noisy or missing) relative 3D distance with respect to the observing agent, together with a one-hot identifier indicating whether the entity corresponds to itself, another agent, or a target. These vectors are embedded and treated as graph vertices; self-attention is applied to learn a latent local coordination graph. An additional recurrent embedding is processed jointly by the Transformer, and the final action distribution is produced by an MLP applied to the transformed recurrent embedding. The TransformerCritic follows the same architecture but is used only during training and receives the full global state (true positions, velocities, and orientations of all agents and targets). It also maintains a recurrent embedding, from which a final MLP outputs the value estimate $V(s)$.}}
      \label{fig:transfmappo}
      \vspace{-0.5em}
   \end{figure}

\textcolor{black}{The complexity of training multi-agent systems with RL typically scales with the number of agents involved. To address this, we introduce \textit{TransfMAPPO}, inspired by TransfQMix~\cite{transfqmix}. Our approach preserves the core ideas of the latter but replaces DQN with PPO as the RL backbone. PPO is naturally suited to vectorised environments, and its clipping mechanism acts as an implicit KL regulariser, which is important for stabilising curriculum learning. To leverage Transformers, we formalise the problem as learning a latent coordination graph through self-attention, where only the vertices of the graph are explicitly defined. Specifically, the $n$ vertices (or tokens) correspond to the agents and targets present in the environment, each represented by $z$ features such as position, velocity, and orientation. This representation is permutation-invariant, allowing the Transformer to learn a coordination graph that generalises across different numbers of agents and targets—much like an LLM can learn semantic attention patterns over sentences of varying length. This structural property of the Transformer, namely its invariance to input ordering and flexibility with respect to input size, makes the network naturally transferable across team sizes, thereby enabling the construction of a curriculum with progressively increasing numbers of agents.}   We distinguish between two types of latent representations: (1) the \textit{local latent graphs} which can be learned by each agent aggregating noisy and potentially incomplete information from onboard sensors and inter-agent communications, and (2) the \textit{global latent graph} learnable using the absolute states of all agents and targets, which is available only during training in simulation. Accordingly, we use two Transformer networks \autoref{fig:transfmappo}: the \emph{TransformerAgents}, deployed in a decentralized fashion (one per vehicle), which learns the local coordination graph to control the agents using only their local observations; and the \emph{TransformerCritic}, which learns a value function for the joint action–state pair using the global information, and is employed for centralized training. The use of Transformers in both actor and critic improves coordination representation and yields policies and value functions that are invariant to the number of entities. This property enables the curriculum learning procedure described in \autoref{sec:curr_learning}.

Following \cite{Gallici25simplifying}, we found that normalizing the entire network with LayerNormalization greatly improves stability and performances.

\vspace{-1em}

\vspace{-0em}
\subsubsection{Curriculum Learning}
\label{sec:curr_learning}

\begin{figure}[thpb]
      \centering
      \includegraphics[scale=0.5]{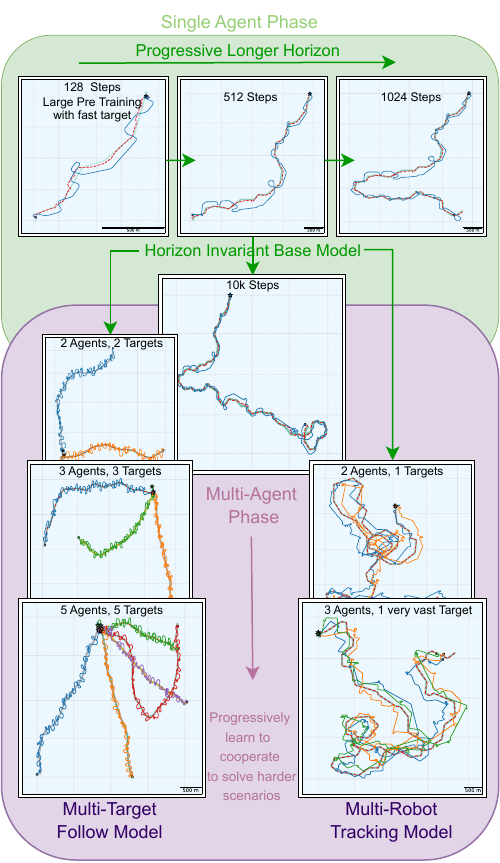}
      \caption{\small Overview of our Curriculum Learning Procedure.}
      \label{fig:curr_learning}
      \vspace{-0.1em}
      \vspace{-2em}
   \end{figure}

Taking advantage of our Transformer-based architecture, we employ curriculum learning to progressively learn Multi-Agent policies starting from a Single-Agent policy (see \autoref{fig:curr_learning}). We begin with a large pre-training phase in which we train a single agent to maximize the \textbf{tracking} reward described in \autoref{sec:environment} with respect to an erratic-moving fast target (0.6x of its speed). We train for $10^{10}$ timesteps, corresponding to 1.5 days of compute (that would take months in Gazebo). In this phase, we maintain a short horizon (128 steps) before resetting the environment, as it significantly aids learning. The resulting policy is able to track a fast target with an error below 35m, but it fails to generalize to longer horizons. We therefore fine-tune the policy in progressively longer episodes (256, 512, and 1024 steps) for $10^{8}$ timesteps, until we observe that the policy can perfectly track a fast target for more than 10k steps, corresponding to more than 3 days of real-time tracking. 

\textcolor{black}{From this horizon-invariant base model, we transition to a multi-agent phase, where we fine-tune the policy for cooperative tasks. The actor parameters are preserved from the single-agent phase and shared across all agents; adding a new agent simply corresponds to instantiating another copy of the same TransformerAgent with identical weights. In particular, we consider two scenarios: (1) $n$ agents tracking $n$ targets, using the \textbf{follow} reward, and (2) $n$ agents tracking one very fast target (0.8x of their velocity), maximising the same \textbf{tracking} reward used in the single-agent phase. When transitioning from the single-agent to the multi-agent setting, we reset the \emph{critic parameters once}, since the value function must now estimate joint returns under coordination and credit assignment effects that were absent in the single-agent case. We then maintain separate multi-agent fine-tuning branches (one per scenario), without further parameter resets as the team size increases. In both cases, the episode horizon remains fixed during this phase, and we fine-tune the model to cooperate with progressively more agents (5 and 3 at maximum, respectively) for a maximum of $2^{9}$ timesteps.}

% We observe that both resulting models can cooperate with a variable number of agents. In this phase, we also note that maintaining long environment horizons negatively affects training, so we reduce the horizon back to 256 steps. Unfortunately, while the resulting multi-agent tracking model retains the horizon invariance of the pre-trained single-agent model, the multi-target follow model loses some invariance in comparison. 

\vspace{-1.2em}

\section{EXPERIMENTS}
\vspace{-0.1em}
All the training was performed on a single H100 GPU of MareNostrum 5 %\footnote{\href{https://www.bsc.es/ca/marenostrum/marenostrum-5}{https://www.bsc.es/ca/marenostrum/marenostrum-5}},
while the Gazebo simulator experiments run on a Intel Sapphire Rapids 8460Y CPU.  All the episode returns are normalized for the horizon length to be presented in the interval $[0,1]$. All the training results are averaged across 5 seeds, except for the curriculum models. All the figures representing agents trajectories are produced with data collected in the Gazebo simulator.

\subsection{Multi-Robot Tracking}

\begin{figure}[thpb]
      \centering
      \includegraphics[scale=0.4]{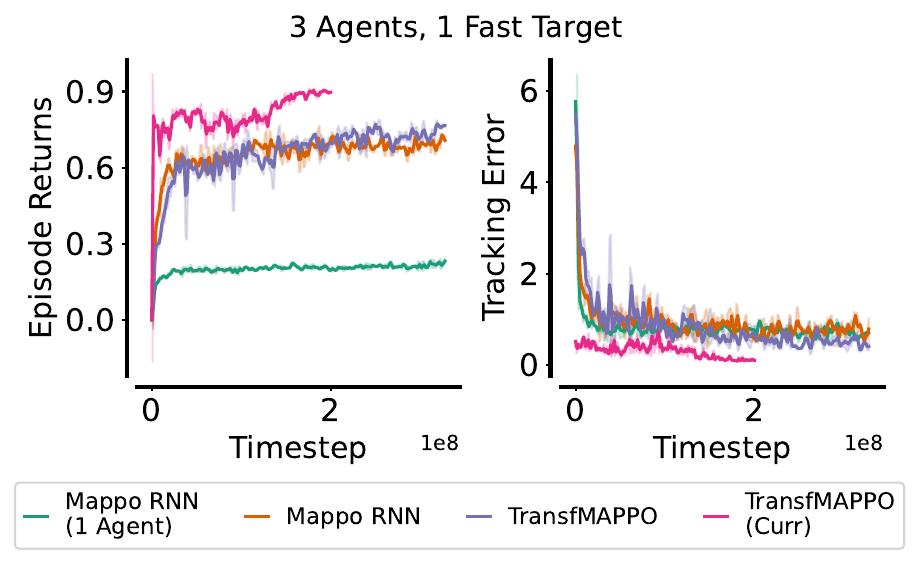}
      \caption{\small Training multiple agents to track a very fast target.}
      \label{fig:multi_robot_training}
      \vspace{-1em}
\end{figure}

In \autoref{fig:multi_robot_training}, we show the training curve for the most complex multi-robot tracking task we consider, i.e., three agents tracking a very fast target (0.8× the agents’ velocity). Specifically, we present results for training standard MAPPO (with an RNN) and TransfMAPPO from scratch, as well as the result of fine-tuning TransfMAPPO on this task following the curriculum procedure described in \autoref{sec:curr_learning}. Note that training Transformers from scratch on this task is not more effective than using an RNN; however, the benefits of curriculum learning are evident in terms of expected returns and reduced tracking error.

\begin{figure}[thpb]
      \centering
      \includegraphics[scale=0.24]{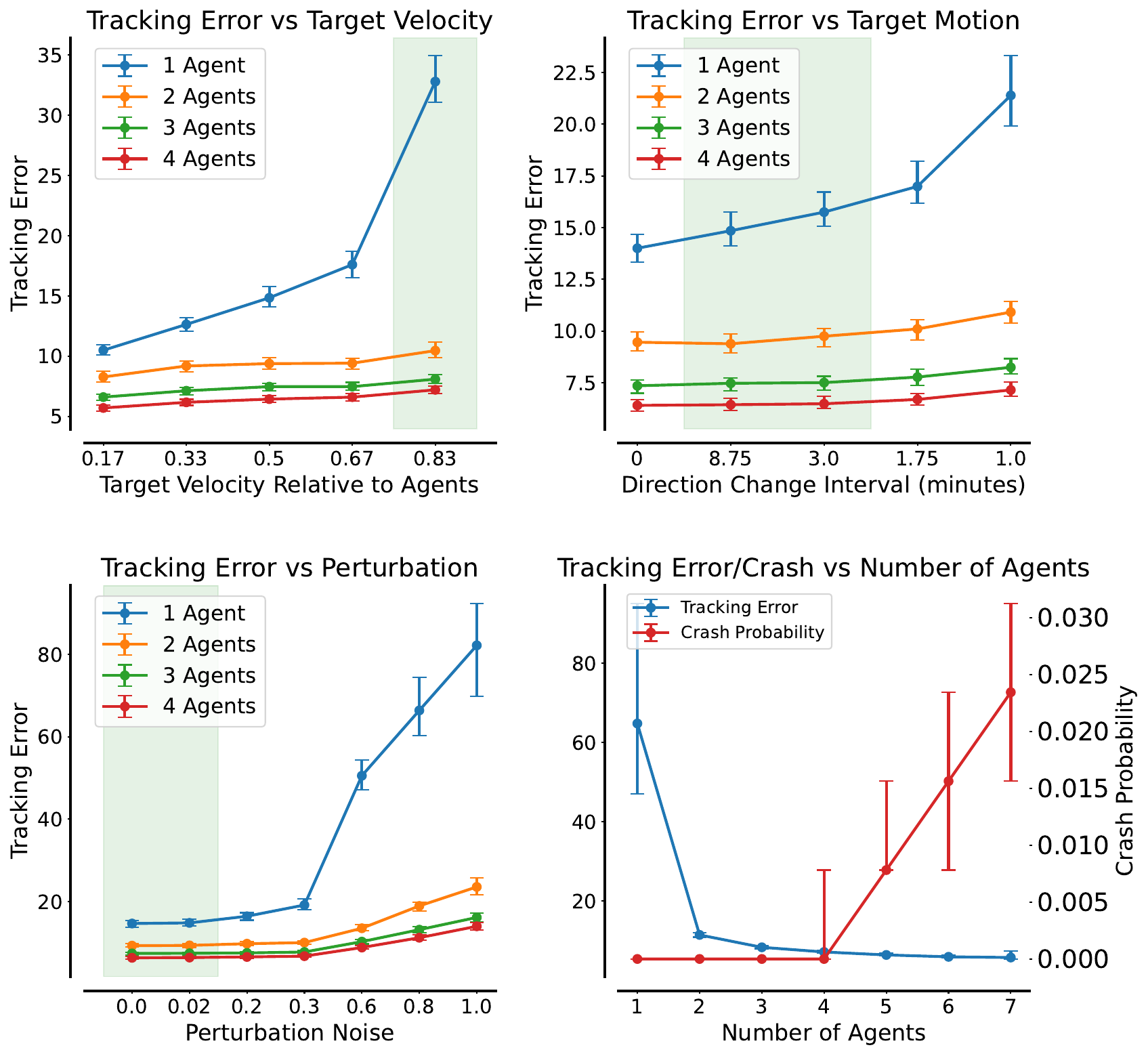}
      \caption{\small Multi-Robot Tracking Evaluation. The highlighted area represents the values seen in training.}
      \label{fig:multi_robot_tracking_test}
      \vspace{-2em}
\end{figure}

In \autoref{fig:multi_robot_tracking_test}, we perform a thorough evaluation of the final multi-target follow model obtained through curriculum training. Each data point in the validation phase is averaged over 1000 episodes, each lasting 256 steps, with the target moving at 0.5× the agents’ velocity (unless otherwise specified). Specifically, we assess how the tracking error is affected under three conditions: (1) when the target moves faster, (2) when its motion becomes more unpredictable (defined by the frequency of direction changes), and (3) when the agents’ motion is perturbed (simulating external forces such as ocean currents). In the last sub-figure, we illustrate how the performance of the multi-agent system scales with the number of agents used to track the target. As expected, even using only two agents significantly improves the system’s robustness compared to a single-agent approach. Robustness increases proportionally with the number of agents, both in terms of resistance to variations in target velocity, target motion (i.e., the interval of target direction changes), and perturbation noise (additional noise added to agents’ trajectories). We also highlight that the tracking error decreases monotonically with the number of agents, with three agents representing a sweet spot: achieving negligible collision probability while maintaining optimal tracking performance.

\subsection{Multi-Target Tracking}

\begin{figure}[thpb]
      \centering
      \includegraphics[scale=0.4]{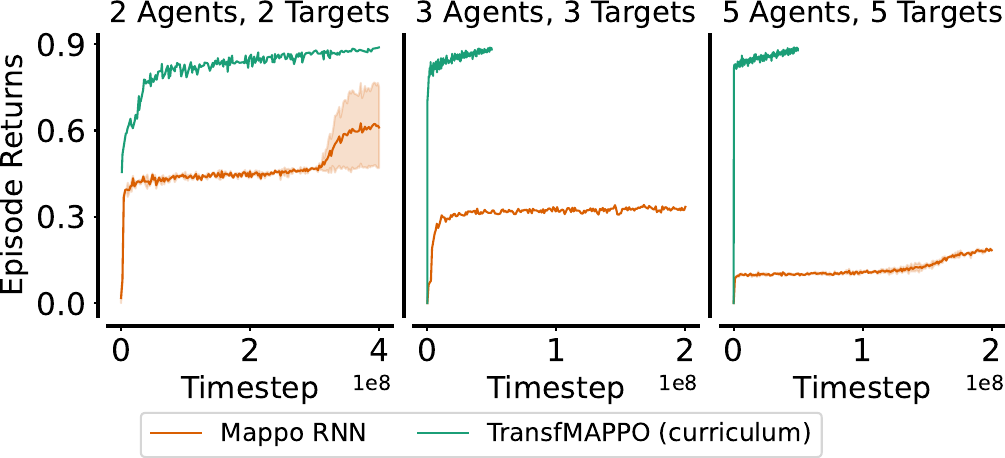}
      \caption{\small Training Multiple Agents to Follow Fast Targets.}
      \label{fig:nvn}
      \vspace{-1em}
\end{figure}

Similarly, our curriculum method achieves remarkable advantages compared to training MAPPO from scratch in the multi-target tracking scenario. In \autoref{fig:nvn}, we present the training curves of training MAPPO from scratch to track moderately fast targets (up to 0.5x the agents' velocity). MAPPO can learn to follow barely half of the targets only in the 2-agent-2-target scenario, while our curriculum approach successfully follows more than 90\% of the targets on average, even in the 5-agent-5-target scenario.

\begin{figure}[thpb]
      \centering
      \includegraphics[scale=0.4]{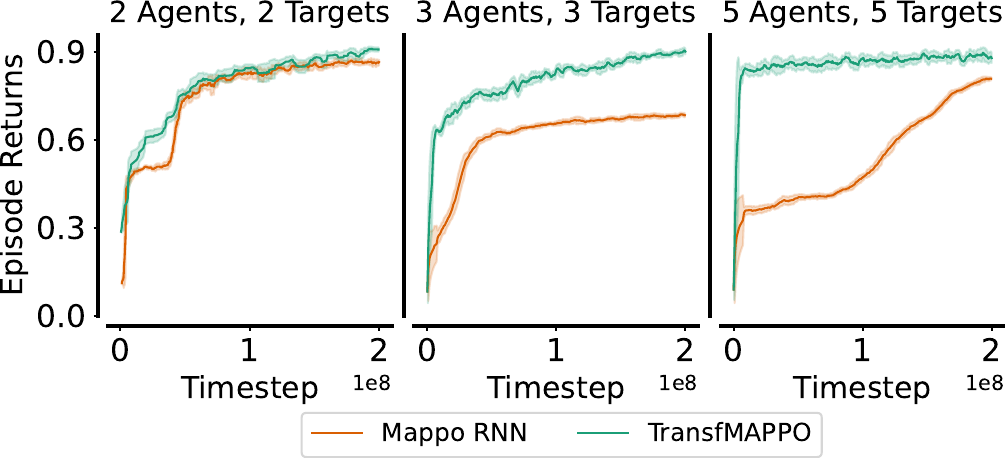}
      \caption{\small Training Multiple Agents to Follow Slower Targets.}
      \label{fig:nvn_easier}
      \vspace{-1.5em}
\end{figure}

To further test the effectiveness of TransfMAPPO, we train it from scratch alongside MAPPO in a simpler configuration where targets move slower (0.3x the agents's speed) and in a more predictable manner (\autoref{fig:nvn_easier}). In this case, TransfMAPPO and MAPPO achieve similar scores with 2 agents, but MAPPO's performance drops significantly with 3 agents. Surprisingly, TransfMAPPO learns very effectively even from scratch in the 5-agent-5-target scenario, confirming the sample efficiency obtained by learning a coordination graph using Transformers.

\subsection{Results in Gazebo}

\begin{table*}[h] 
\vspace{0.5em}
\caption{\small Evaluation in JaxLrauv and PyLrauv of the curriculum training policies (* except for first line which refers to 10 minutes training).}
\resizebox{\textwidth}{!}{%
\begin{tabular}{l 
                c c 
                c c 
                c c 
                c c}
\toprule
\multirow{2}{*}{Configuration} 
& \multicolumn{2}{c}{Avg. Agent-Target Distance} 
& \multicolumn{2}{c}{Avg. Tracking Error} 
& \multicolumn{2}{c}{Probability of Collision (\%)} 
& \multicolumn{2}{c}{Probability of Loosing a Target (\%)} \\
\cmidrule(lr){2-3}
\cmidrule(lr){4-5}
\cmidrule(lr){6-7}
\cmidrule(lr){8-9}
& JaxLrauv & PyLrauv 
& JaxLrauv & PyLrauv 
& JaxLrauv & PyLrauv 
& JaxLrauv & PyLrauv \\
\midrule
1 Agent, 1 Target (Slow)\textbf{*}
& 55.20 ± 30.11 & 59.10 ± 32.29 
& 5.12 ± 3.1 & 5.89 ± 3.7 
& 0.00 & 0.00 
& 0.00 & 0.00 \\
1 Agent, 1 Target (Fast) 
& 110.23 ± 58.90 & 121.01 ± 62.11 
& 17.40 ± 21.5 & 20.33 ± 28.38 
& 0.00 & 0.00 
& 5.00 & 7.14 \\
3 Agents, 1 Target (Very Fast) 
& 200.11 ± 85.70 & 217.64 ± 89.85 
& 2.65 ± 5.30 & 3.03 ± 6.17 
& 0.0 & 15.32 
& 0.00 & 0.00 \\
3 Agents, 3 Targets (Moderate) 
& 42.15 ± 25.00 & 45.30 ± 26.11 
& 4.85 ± 4.20 & 5.29 ± 4.88 
& 0.5 & 5.00 
& 3.50 & 5.00 \\
5 Agents, 5 Targets (Moderate) 
& 46.12 ± 28.75 & 49.24 ± 30.92 
& 3.80 ± 4.00 & 4.25 ± 4.57 
& 2.1 & 10.00 
& 4.10 & 5.26 \\
\bottomrule
\end{tabular}%
}
\label{tab:gazebo-results}
\vspace{-1em}
\end{table*}

We conducted a series of final experiments in Gazebo to evaluate the performance of our trained agents in a high-fidelity simulation environment. We collected data over 50 episodes using the models obtained at the conclusion of the curriculum training. Specifically, we performed the following tests:

\begin{enumerate}
    \item Tracking a slow target (0.3$\times$ the agent's velocity) for 300 steps using a single robot trained for 10 minutes in JaxLrauv.
    \item Tracking a fast target (0.6$\times$ the agent's velocity) for 1000 steps (equivalent to over 8 hours of real-time tracking) using our horizon-invariant base model.
    \item Tracking a very fast target (0.8$\times$ the agent's velocity) for 1000 steps using a multi-robot tracking fine-tuned model with 3 agents.
    \item Tracking 3 targets simultaneously moving at 0.5$\times$ the agents' velocity using a multi-target model for 300 steps.
    \item Tracking 5 targets simultaneously moving at 0.5$\times$ the agents' velocity using a multi-target model for 300 steps.
\end{enumerate}

\begin{figure}[thpb]
  \centering
  \subfloat[]{%
    \includegraphics[width=0.32\columnwidth]{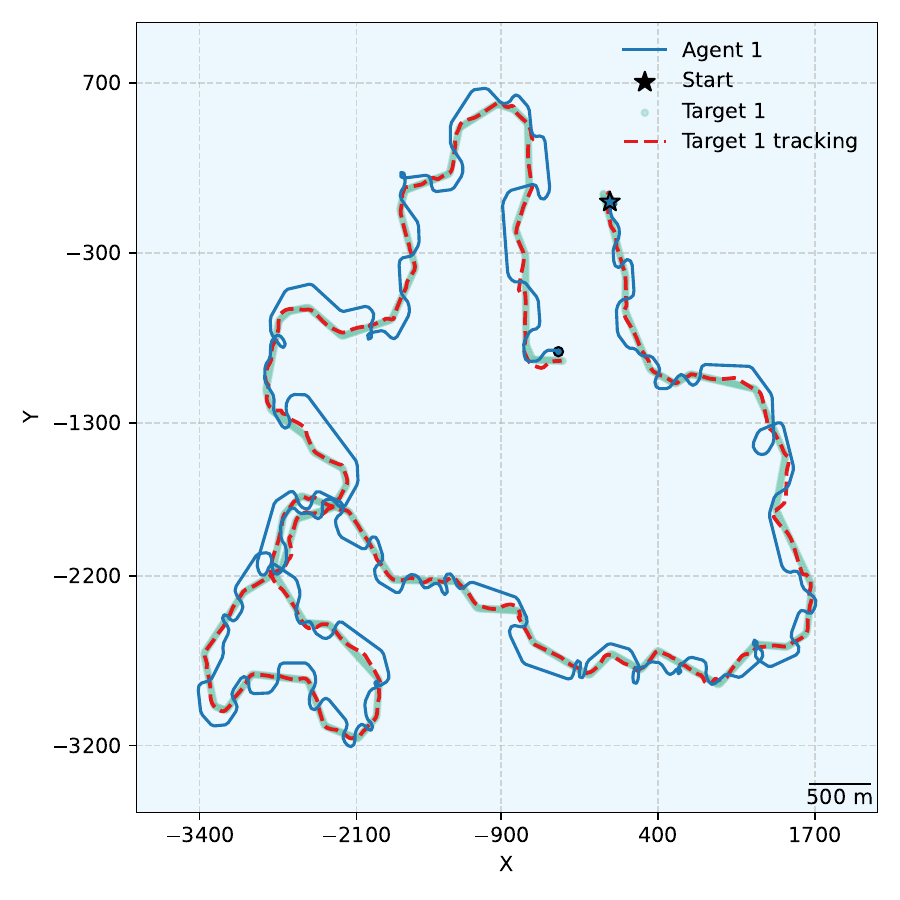}%
    \label{fig:1target_test_a}%
  }%
  \hspace*{2em}% 
  \subfloat[]{%
    \includegraphics[width=0.32\columnwidth]{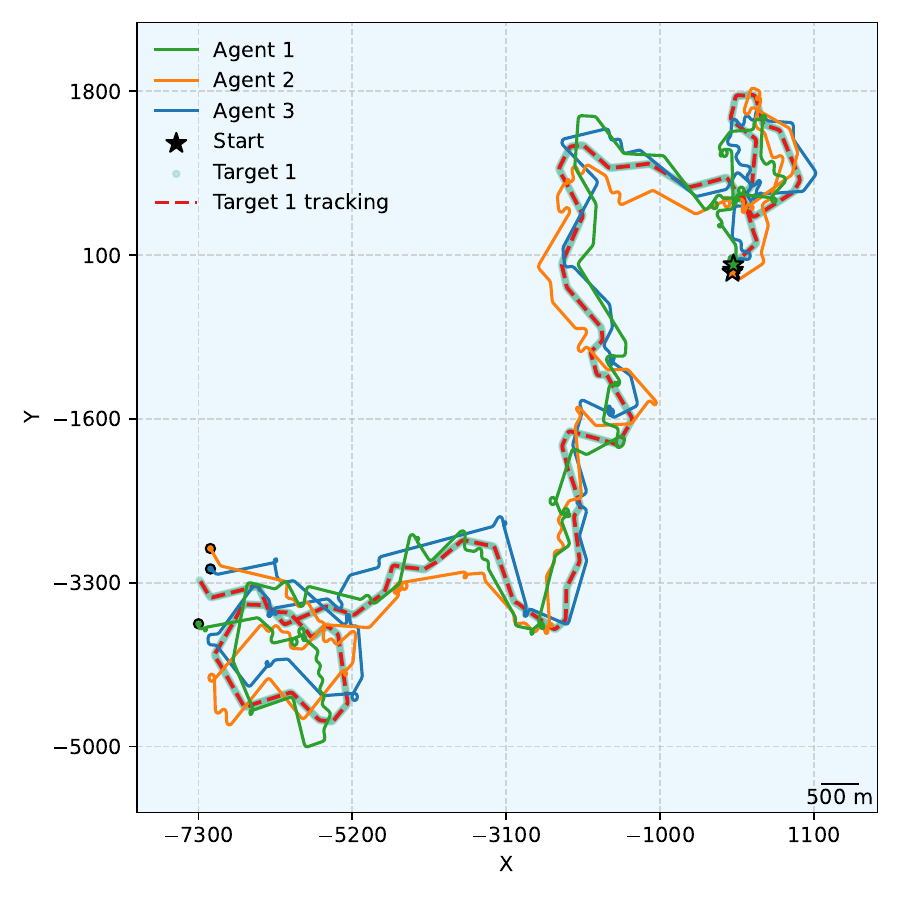}%
    \label{fig:1target_test_b}%
  }%
  \vspace{-0.5em}
  \caption{\small (a): Single-agent tracking, target velocity 0.66× of agent velocity. \href{https://mttga.github.io/posts/pylrauv/images/1v1.gif}{Video.} (b): Multi-agent tracking, target velocity 0.88x of agents' velocity. \href{https://mttga.github.io/posts/pylrauv/images/3v1.gif}{Video.}}%
  \label{fig:1target_test}%
  %\vspace{-0.5em}
\end{figure}

\begin{figure}[thpb]
      \centering
      \includegraphics[scale=0.2]{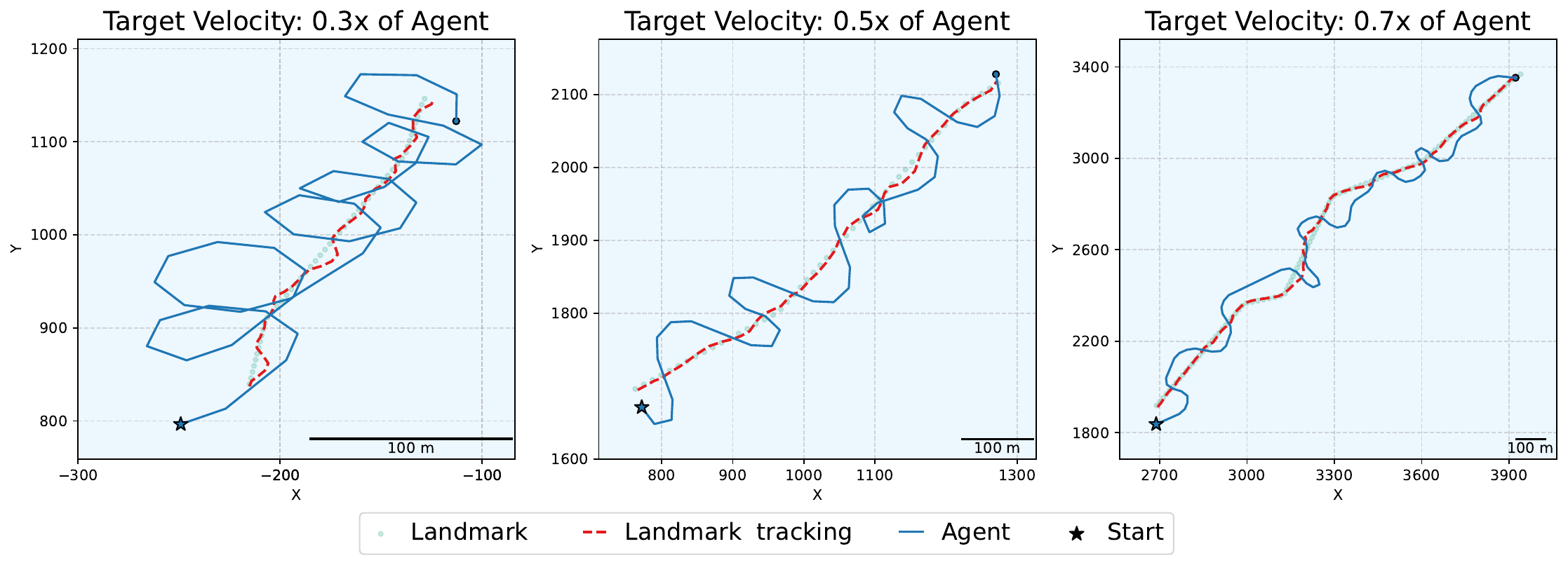}
      %\vspace{-0.8em}
      \caption{\small Agents' trajectory according to target speed. Notice the change in the curvature of the agent's trajectory to optimally track targets moving at different speeds.}
      \label{fig:motion_vs_speed}
      %\vspace{-1em}
\end{figure}

\begin{figure}[!h]
      \centering
      \includegraphics[scale=0.2]{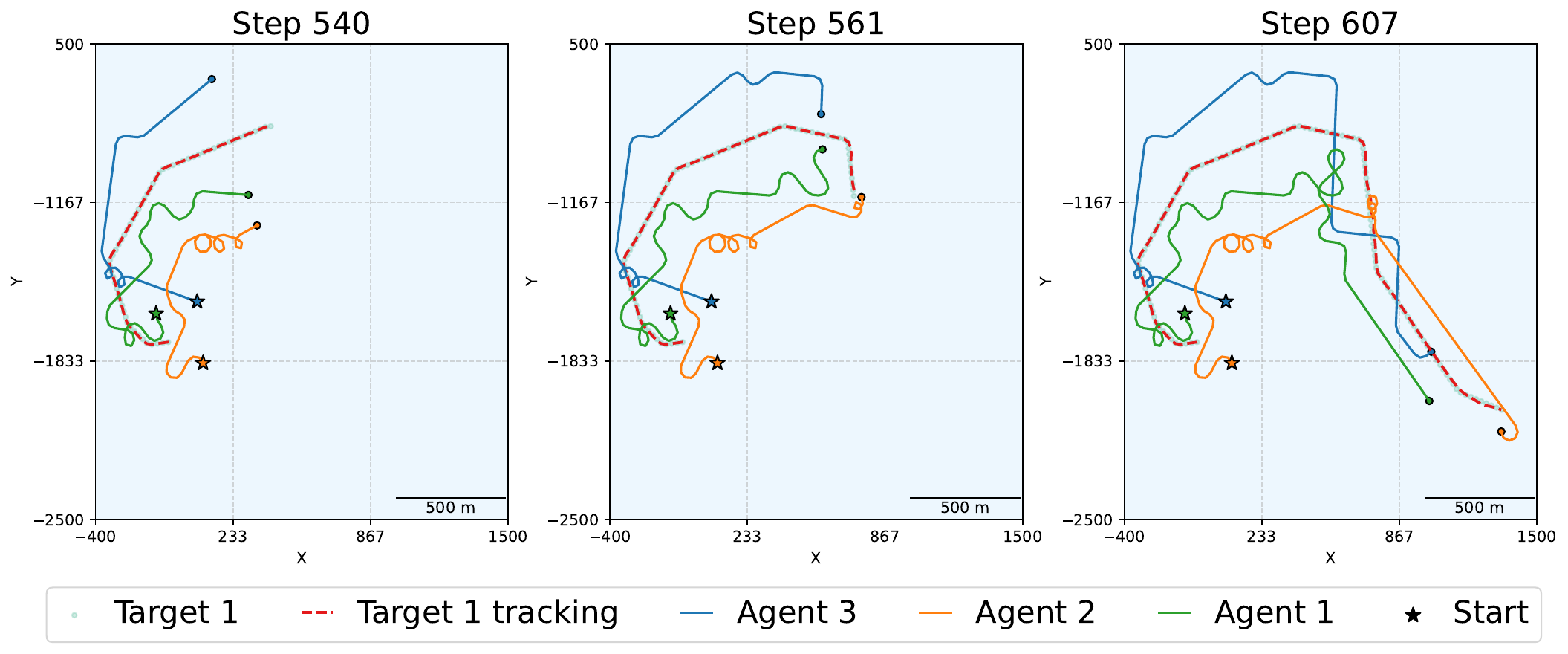}
      %\vspace{-0.5em}
      \caption{\small Coordination when tracking a very fast target. Notice how agents react to a rapid change of direction of the target by waiting for one other in order not to lose communication and then rapidly moving together.}
      \label{fig:3v1_progressive}
      %\vspace{-1em}
\end{figure}

\begin{figure}[!h]
      \centering
      \includegraphics[scale=0.2]{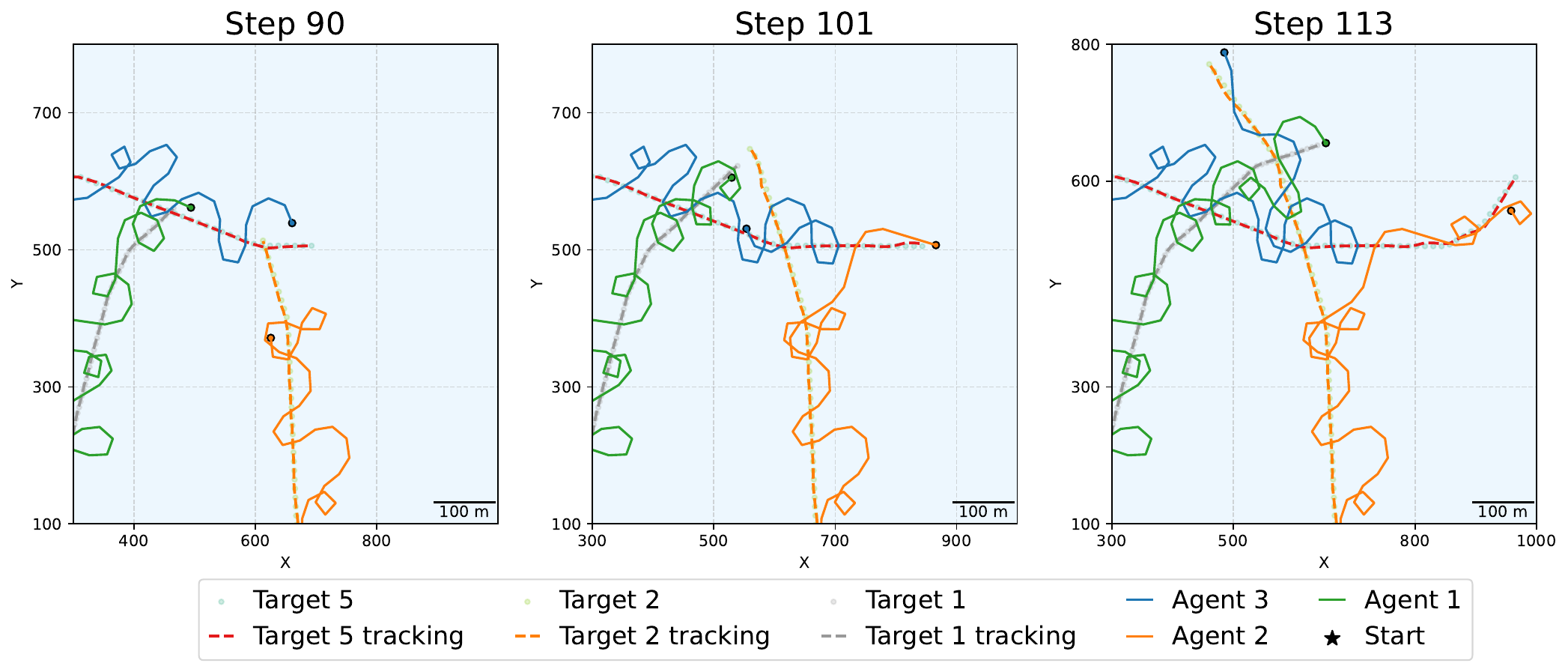}
      %\vspace{-0.5em}
      \caption{\small  Coordination in a multi-target setting. Notice how blue and orange agents resolve the "traffic" problem by making circles to wait for other agents' passing, and one target is dynamically reassigned from one agent to another.}
      \label{fig:5v5_progressive}
      %\vspace{-1em}
\end{figure}

For the multi-target experiments, we reduced the number of steps to 300 due to the computational burden of simulating multiple robots in Gazebo. The results of these experiments are presented in \autoref{tab:gazebo-results}. For completeness, we also include in this table the results obtained in JaxLrauv with the same configurations, so that it is possible to appreciate how the learned policy behaves in a very similar way in both simulators, resulting in comparable evaluation measures. Our findings demonstrate that even with only 10 minutes of training, we were able to develop an effective model capable of maintaining target tracking without losing the target. The multi-robot tracking model achieved a promising average tracking error of 3m, even when following a very fast target. This performance highlights a clear advantage over single-agent trackers following slower targets. Notably, the average agent-target distance increased when using multiple robots, reflecting a broader distribution of robots around the target space. The multi-target models performed as expected, with a small 5\% probability of losing a single landmark. However, the multi-agent models exhibited an increased probability of collisions. This issue is likely attributable to the fact that the models were trained to avoid collisions only every 30 seconds, which can result in intra-step collisions in Gazebo. This underscores a critical area for future work to enhance the systems' safety before deploying them in real-world marine missions. Finally, in \Cref{fig:1target_test,fig:motion_vs_speed,fig:3v1_progressive,fig:5v5_progressive}, we present some of the trajectories obtained during the final evaluation, together with zoomed-in views of the emergent coordination strategies learned. 

%\vspace{-2em}

\vspace{-1em}

%\vspace{-0.5em}
\section{CONCLUSIONS}

We presented a method for scaling up Multi-Agent Reinforcement Learning (MARL) in the context of underwater tracking, demonstrating how it is possible to efficiently train MARL models purely on GPUs and deploy them in high-fidelity simulators, where MARL training would otherwise be extremely expensive or infeasible. Our overall pipeline represents a promising approach for closing the sim-to-real gap in MARL. A natural next step for this work is to enhance the safety of the multi-agent system and test it in real-world sea missions.

%\scriptsize
\section*{ACKNOWLEDGMENT}
This work acknowledges the Spanish Ministerio de Ciencia, Innovación y Universidades (TECTUGA: PID2024-161772OA-I00). This work is part of DIGI4ECO, European Union’s Horizon Europe programme (No 101112883). I. M. received financial support from the MCIN/AEI/10.13039/501100011033 and FSE+ (No RYC2022-038056-I). M. G. was partially funded by the FPI-UPC Santander Scholarship FPI-UPC\_93. This work also acknowledges the ‘Severo Ochoa Centre of Excellence’ accreditation (CEX2024-001494-S) from AEI 10.13039/501100011033 and the computing resources provided by the Barcelona Supercomputing Center (BSC).

%%%%%%%%%%%%%%%%%%%%%%%%%%%%%%%%%%%%%%%%%%%%%%%%%%%%%%%%%%%%%%%%%%%%%%%%%%%%%%%%

%\addtolength{\textheight}{-1cm}    % This command serves to balance the column lengths
                                  % on the last page of the document manually. It shortens
                                  % the textheight of the last page by a suitable amount.
                                  % This command does not take effect until the next page
                                  % so it should come on the page before the last. Make
                                  % sure that you do not shorten the textheight too much.

\balance

\AtNextBibliography{\small}
\printbibliography

%%%%%%%%%%%%%%%%%%%%%%%%%%%%%%%%%%%%%%%%%%%%%%%%%%%%%%%%%%%%%%%%%%%%%%%%%%%%%%%%

%%%%%%%%%%%%%%%%%%%%%%%%%%%%%%%%%%%%%%%%%%%%%%%%%%%%%%%%%%%%%%%%%%%%%%%%%%%%%%%%

%%%%%%%%%%%%%%%%%%%%%%%%%%%%%%%%%%%%%%%%%%%%%%%%%%%%%%%%%%%%%%%%%%%%%%%%%%%%%%%%

\end{document}